\documentclass[letterpaper, 10 pt, conference]{ieeeconf}

\IEEEoverridecommandlockouts                       
\usepackage[english]{babel}
\usepackage{amsmath}
\usepackage{amssymb}
\usepackage{booktabs}
\usepackage[dvips]{graphicx}
\usepackage{subfigure}
\usepackage{multirow}
\usepackage{caption}
\usepackage{relsize}
\usepackage{amsfonts}
\usepackage{tabularx}
\usepackage{amsfonts}
\usepackage{algorithm,algorithmic}
\usepackage{bm}
\usepackage{enumerate}
\usepackage{siunitx}
\usepackage{adjustbox}
\usepackage{authblk}
\usepackage{color}
\usepackage{xurl}
\usepackage{scalerel}
\usepackage{cite}
\makeatletter
\let\NAT@parse\undefined
\makeatother
\usepackage{hyperref}

\ExplSyntaxOn
\NewDocumentCommand{\rvect}{m}
 {
  \seq_set_split:Nnn \l_tmpa_seq { , } { #1 }
  \begin{bmatrix}
  \seq_use:Nn \l_tmpa_seq { & }
  \end{bmatrix}
 }
\ExplSyntaxOff

\makeatother

\title{\LARGE \bf
Spatial Assisted Human-Drone Collaborative Navigation and Interaction through Immersive Mixed Reality}

\author{Luca Morando and Giuseppe Loianno 
\thanks{The authors are with the New York University, Tandon School of Engineering, Brooklyn, NY 11201, USA. \tt\footnotesize email: \{luca.morando, loiannog\}@nyu.edu.}
\thanks{This work was supported by the NSF CPS Grant CNS-2121391, the NSF CAREER Award 2145277, the DARPA YFA Grant D22AP00156-00, Qualcomm Research, Nokia, and NYU Wireless.}
}

\begin{document}
\thispagestyle{empty}
\pagestyle{empty}
\makeatletter

\g@addto@macro\@maketitle{
    \setcounter{figure}{0}
    \centering
\includegraphics[width=\textwidth, keepaspectratio]
    {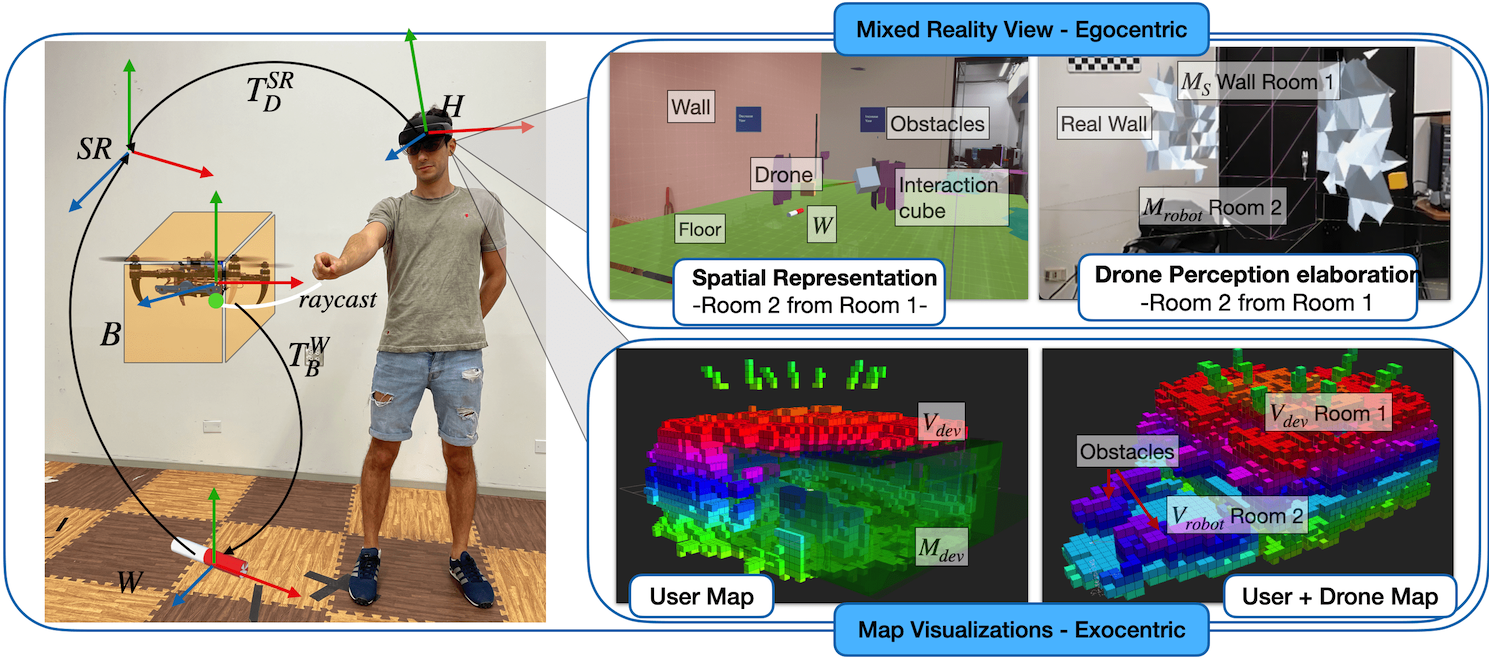}
    \captionof{figure}{ a.) \emph{Human-Drone interaction}: Pinch and drag gestures are available to manipulate and to interact with the hologram of the robot. b.) \emph{Spatial-Awareness}: the robot flying space (left image) and the drone generated mesh (right image) are spatially represented with a through-the-wall visualization, c.) \emph{Maps Merging}: visualization of the merged user and drone maps of the two different rooms.}
    \label{fig:fig1}
    \vspace{-15pt}
}

\makeatother
\maketitle


\begin{abstract}
Aerial robots have the potential to play a crucial role in assisting humans with complex and dangerous tasks. Nevertheless, the future industry demands innovative solutions to streamline the interaction process between humans and drones to enable seamless collaboration and efficient co-working. In this paper, we present a novel tele-immersive framework that promotes cognitive and physical collaboration between humans and robots through Mixed Reality (MR). This framework incorporates a novel bi-directional spatial awareness and a multi-modal virtual-physical interaction approaches. The former seamlessly integrates the physical and virtual worlds, offering bidirectional egocentric and exocentric environmental representations. The latter, leveraging the proposed spatial representation, further enhances the collaboration combining a robot planning algorithm for obstacle avoidance with a variable admittance control. This allows users to issue commands based on virtual forces while maintaining compatibility with the environment map. We validate the proposed approach by performing several collaborative planning and exploration tasks involving a drone and an user equipped with a MR headset.
\vspace{-5pt}
\end{abstract}

\section*{Supplementary Material}
\noindent\textbf{Code:} \url{https://github.com/arplaboratory/mri_ros_public.git}
\noindent\textbf{Video:} \url{https://youtu.be/oWdqLuXe1Zw}

\section{Introduction}
The increased use of robotics systems in our everyday life raises the need to design novel solutions that can facilitate the interaction between humans and robots decreasing the physical and cognitive human workload especially when executing complex or dangerous tasks. Human-Robot Collaboration and Interaction (HRI) is a well-studied problem and being integrated in multiple areas, such as Industry 4.0, construction, and emergency responses~\cite{Ajoudani2018,Morando2020}.
Traditionally, human-centered robotic systems focus on collaboration between humans and grounded robots.
In aerial robotics, due to the additional complexity introduced by the navigation requirements in 3D unstructured environments, current state of the art approaches still limit the collaboration to be mostly unidirectional (i.e., with commands or information sent from the human to the robot) using classic teleoperation~\cite{SehoonHRD} with representation of the output signals projected on flat surface monitors.
Aerial robotics will become essential in supporting humans in multiple tasks including, but not limited to warehouses' operations, vertical farming inspection, space missions, and search and rescue. In these settings, robots' role can no longer be limited to simply assist humans. Instead, they will need to co-exist, collaborate, and co-work with them, sharing the same workspace and participating in shared tasks' execution. Therefore, these requirements  call for the novel approaches that enable a high level of mutual and seamless human-robot interaction. 
We envision aerial robots to be elevated to the role of humans' teammates allowing a bi-directional flux of information to concurrently augment both human and robot situational awareness.

In this work, we propose a novel tele-immersive framework for human-drone cognitive and physical collaboration through Mixed Reality (MR) as illustrated in Fig.~\ref{fig:fig1} to facilitate seamless interaction between humans and robots.
The emergence of innovative spatial computing techniques, such as Virtual Reality (VR), Augmented Reality (AR), and Mixed Reality (MR), presents a technological opportunity in robotics. These techniques facilitate enhanced collaboration between humans and robots through multi-modal information sharing within the human-robot team based on vision~\cite{HALME2018111}, gestures~\cite{LIU2018355}, natural languages~\cite{Pate2021}, and gaze~\cite{LoiannoIcra,Pavliv2021}.

The contributions of this paper can be summarized as follows.
First, we present a novel bi-directional spatial awareness concept that enables co-sharing of spatial information between a human and an aerial robot at different levels of abstractions, making them co-aware of the surrounding environment. 
Second, we propose a novel virtual-physical interaction that further facilitates bi-directional collaboration. This approach uniquely combines a  Variable Admittance Control (VAC)~\cite{LecoursIcra2012,FerragutiIJRR2019}, implemented on top of the drone's guidance loop,  with a  planning algorithm \cite{WangAppliedSciences2020} responsible for robot's obstacle avoidance. The proposed VAC allows through gesture recognition to input a user virtual external force as an external command to the robot while ensuring compatibility with the environment map.
The user force is further combined with a repulsive force field acting on the robot when located in proximity of obstacles increasing the overall system safety.

To the best of our knowledge, the proposed framework is the first to enable continuous spatial virtual-physical navigation and interaction with an aerial robot via  MR. We open source the framework to the community.


\section{Related Works}~\label{sec:related_works}
\textbf{Teleoperation}. Drone teleoperation, while arguably one of the simplest interaction methods, remains an open research area especially in constrained robot environments~\cite{Isop2019,Rojas2019}. 
Current approaches still rely on simple joysticks, keyboards, and pointers receiving the feedback from the robots on monitors. 
Overall, state of the art autopilots can enable drones to follow aggressive trajectories  \cite{Ales2022}, 
but they do not offer an easy and intuitive communication paradigm between human and robot.
Recent solutions try to overcome these drawbacks by employing hand gesture~\cite{nagi2014human} or impedance control with vibrotactile feedback \cite{tsykunov2019swarmtouch} to control multiple drones or impedance control for human-machine~\cite{Tagliabue2017,Augugliaro2013} or  machine-environment interaction~\cite{Car2018}.

\textbf{Extended Reality}. Innovative body-machine interfaces or wearable devices (e.g., head mounted displays) utilize body motion to control robots and incorporate haptic feedback, along with VR or AR~\cite{macchini2019personalized,VaughanCRV2017,nagi2014HRI,walker2018communicating,reardon2018ssrr}. These interfaces aim to address the constraints of basic teleoperation control. For instance, AR proves to be a valuable interaction modality that facilitates collaborative information visualization in robotics tasks, therefore enabling efficient communication of robots' intentions to human co-workers~\cite{coovert2014spatial,rosen2017communicating}.
In \cite{Duncan_gestures}, the authors study how to maximize the user ability to tele-operate a particular drone projecting its camera field of view camera in the user headset.
In \cite{Erat2018,Shaojie2020,Betancourt2022}, an exocentric view of the drone pose in a virtual scene that is built a-priori, helps the user to interact with the drone via a pick and place gesture, commanding a new goal position to the robot. 
A gesture based trajectory completion in MR is proposed in \cite{Zein2021}, where an automated technique auto-complete the user intended motion.
Conversely, the authors in \cite{Angelopoulos2022} show a collaborative MR interface for immersive planning where the user is able, using simple gestures to draw 3D trajectories for quadrotors in its workspace. While in \cite{Angelopoulos2022} the main focus is on drone teleoperation via simple gestures, in \cite{Reardon2019} the authors employ MR to analyze the information sharing problem, in order to simplify the human-robot communication during an exploration task.
Information about the robot's future actions and predicted trajectory is visualized using AR. Finally, a recent work \cite{Szafir2} proposes a fascinating instance of enhancing human situational awareness in MR during human-robot interaction. The MR experience merges a downsized satellite map with the robot's local environment representation. Users have the ability to choose a location on the map and stream the locally captured robot's point cloud into the MR headset, providing them with a first-person view of the scene.

\textbf{Novelties}. Our tele-immersive approach goes beyond simple tele-operation and extended reality for robotics. It facilitates a bi-directional flux of information to enhance the spatial awareness of both the human and the drone.  This aspect is further enhanced incorporating a novel human-robot virtual-physical interaction in 3D based on a VAC~\cite{Bazzi2022}. Compared to~\cite{Bazzi2022}, our work goes beyond the manipulator case, it eliminates the need for prior knowledge about the environment's structure and seamlessly and uniquely designs and deploys the VAC within the online 3D navigation framework.

\section{Methodology}~\label{sec:Methodology}
As visible in Fig.~\ref{fig:fig1}.a, the robot body frame is denoted with $B$ and the robot world frame, located at the drone take-off position, is denoted with $W$ and it is represented in the MR environment as a red and white cylinder. Referring to Fig.~\ref{fig:fig1}.a, we denote the spatial transformation between two frames $i$ and $j$ as with $\mathbf{T}_{j}^i$ that represents the relative pose of the $j$ frame with respect to the $i$.


\subsection{Spatial Awareness}~\label{sec:spatial_awareness}

\vspace{-10pt}
\begin{figure*}[t!]
\centering
\vstretch{1}{\includegraphics[width=\textwidth, keepaspectratio]{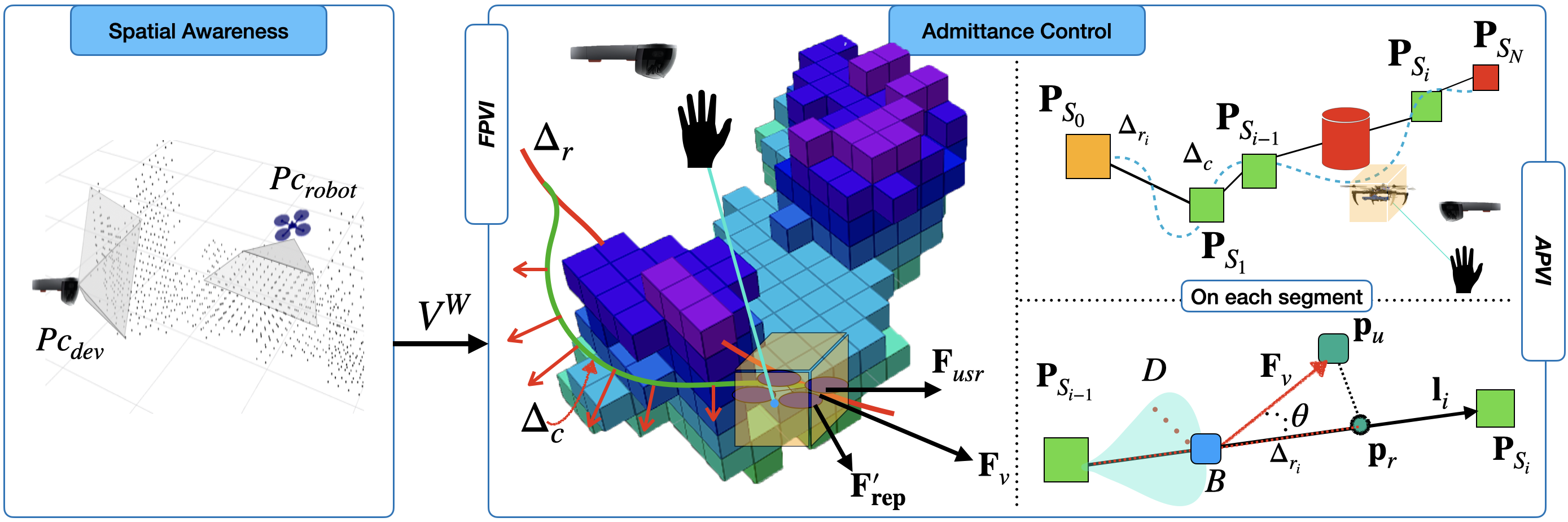}}
  \caption{Key framework's components. The pointcloud set $P^W$ is collected and sent to the SAM where a collective octomap is generated (left). A VAC coupled with a planner provides assistance to the user during the drone navigation (right). }
  \label{fig:fig2}
  \vspace{-20pt}
\end{figure*}

We introduce the concept of human-drone spatial awareness. This cutting-edge technique facilitates an improved representation of the robot's perceived environment on the user side, utilizing advanced mapping and visualization methods. Moreover, on the robot side, the user's perception of the  surrounding environment is translated in a map representation that is interpretable by the robot navigation layer. The raw data obtained from the drone and the wearable device sensors are processed through the Spatial Awareness Module (SAM) as presented in Fig.~\ref{fig:fig1}. 
Specifically, the robot point cloud ${P_c}_{robot}$ (in our case a stereo disparity as detailed in Section~\ref{sec:Experimental}) and the wearable device point cloud ${P_c}_{dev}$ are forwarded as input to the robot SAM which provides as output two types of geometric representations namely an octomap and a mesh visualization shown in Fig.~\ref{fig:fig1}.b and Fig.~\ref{fig:fig1}.c respectively.

The set of the robot and device pointclouds ${P_c}^{W} = \{{P_c}_{dev}, {P_c}_{robot}\}^{W}$ can be defined and expressed in $W$ using the following transformations: i) $\mathbf{T}_{B}^W$ for  ${P_c}_{robot}$ and ii) $\mathbf{T}_{SR}^W$ for ${P_c}_{dev}$, where the frame $SR$ is initialized with the same pose of the device frame $H$, when the MR application is launched.
Leveraging the Voxblox framework \cite{Voxblox2017}, it is possible to obtain  an octomap set from the pointcloud ${P_c}^{W}$, denoted as $V^W=\{V_{robot}, V_{dev}\}^W$, where $V_{robot}$ is the voxel representation on the robot side while $ V_{dev}$ is similarly obtained from the user perspective. The set $V^W$ define a unified global voxel representation visible in Fig.~\ref{fig:fig1}.c where the merged maps of two adjacent rooms (one explored by the robot and the other one explored by the user) are depicted.
An example of the representation instantiated by the SAM is depicted in Fig.\ref{fig:fig1}.b and Fig.\ref{fig:fig1}.c, showcasing a dual visualization of the environments perceived by the user and the robot, respectively.

In Fig.~\ref{fig:fig1}.b, we show the user Ego-centric visualization that allows the MR layer to translate the drone perceived environment in a spatial language easily understandable by the human.
The Exo-centric visualization instead, depicted in Fig.~\ref{fig:fig1}.c, merges the user and the drone maps captured from two different rooms or perspectives, therefore augmenting the drone perception capabilities.   
A further level of abstraction and functionality is provided by the instantiation of the MR environment in the user field of view, where both the wearable device frame $H$ and the world frame $W$, are referred to the $SR$ frame via $\mathbf{T}_{W}^{SR}$ and $\mathbf{T}_{H}^{SR}$ respectively, as shown in Fig.~\ref{fig:fig1}.a.
The wearable device can extract information in the perceived environment about various surfaces (i.e.,  walls, ceilings, floors etc.) or artificially create additional elements in form of holograms around the user (see Fig. \ref{fig:fig1}.b). This produces a seamless mix of real and virtual worlds allowing a customized and reconfigurable spatial interaction with increased configuration complexities between the human and/or the robot even in an empty physical space.  An example is provided in Fig.~\ref{fig:fig1}.b, where the drone mesh $M_{robot}$, obtained by ${P_c}_{robot}$ and visible in white, is merged with other spatially defined holograms in the MR framework. 

Finally, the drone pose is always represented in the MR environment through an orange cube $C$ referred to the frame $W$ (see Fig.~\ref{fig:fig1}.a). 

\subsection{Virtual-Physical Human-Drone Interactive Navigation}~\label{sec:PVH interaction}
We propose a VAC to link the physical and virtual worlds producing two types of virtual-physical  interaction

\begin{itemize}
    \item \textbf{Assisted Physical and Virtual Interaction (APVI)}: In this modality, a planning algorithm (in our case a Rapidly exploring Random Tree in its optimal formulation (RRT*)~\cite{WangAppliedSciences2020}) is built on top of the generated octomap and combined with a VAC algorithm. Concurrently, the VAC provides a modulated damping feedback based on a desired function. This can modify the drone compliance depending the user intention to modify its path from the original one proposed by the planner.
    \item \textbf{Free Physical and Virtual Interaction (FPVI)}: This is a specific case of the APVI modality when the RRT* planner is disabled by the user. This modality provides additional freedom when driving the robot. Since the robot is not anymore constrained to follow a specific path, an additional safety layer is required. We propose the generation of an obstacle force field that is injected in the admittance controller as an additional input.
\end{itemize}

A schematic representation of both modalities with their integration with the SAM is visualized in Fig.~\ref{fig:fig2}.
The two modalities offer varying levels of intuitiveness and environmental safety during the human-robot interaction, allowing the user to depend more on the robot's perception of its surroundings for effective exploration, especially in unfamiliar or obstacle-filled environments. 
In the following, we describe the key components of the APVI and FPVI namely the RRT*, the VAC and how this is couple with the RRT* for the APVI modality, and the obstacle force field.
\subsubsection{RRT* assisted Human-Robot Navigation}\label{sec:RRT* Intro}
The RRT* is enabled in the APVI mode, expanding its tree based on a local robot map defined within a circular horizon centered on the drone body frame origin $\mathcal{O_B}$. 
This avoids the problem to know a-priori the environment map and local minima planning issues can be taken care by the  user intervention with the proposed VAC.
The output of the RRT* is a sequence of setpoints $\mathbf{P}_{S_i}$, for $i=0,\cdots N$, which are computed only within the drone horizon.
The path is updated on-line, while the drone is moving along the proposed direction. As visible in 
Fig.~\ref{fig:fig2} (right), the robot reference trajectory 
is composed by a sequence of reference positions  $\mathbf{p}_r$ that lies on the line each defined by a segment between two consecutive setpoints $\mathbf{P}_{S_i}$ and $\mathbf{P}_{S_{i+1}}$ generated by the RRT*. The sequence of reference position $\mathbf{p}_r$ compose the trajectory $\mathbf{\Delta}_{r}$.

During the random nodes exploration phase embedded in the RRT* , the total cost of the new node $c^*$ is obtained by adding the linking cost $c_{l}$ to the parent cost $c_{p}$. 

\subsubsection{Coupling Planning and VAC for assisted virtual-physical HRI} \label{sec:Coupling Planning VAC}
We combine the RRT* planning algorithm, presented in Section \ref{sec:RRT* Intro} with a VAC.
Prior to describing the VAC, we design an admittance controller on top of the drone inner control position pipeline.
This controller, in its classic formulation visible in eq.~(\ref{eq:admittance1}), is driven by the 3D virtual input force $\mathbf{F}_v$ generating the position perturbation $\mathcal{X}= \mathbf{p}_c - \mathbf{p}_r$ between a commanded position $\mathbf{p}_c$ and a reference position $\mathbf{p}_r$. The dynamical relationship between the input force $\mathbf{F}_v$ and the position perturbation $\mathbf{\mathcal{X}}$ mimics a mass spring damper system as
\begin{equation}
    \mathbf{M} \left(\ddot{\mathbf{p}}_c - \ddot{\mathbf{p}}_r\right)+ \mathbf{D}\left(\dot{\mathbf{p}}_c - \dot{\mathbf{p}}_r\right) + \mathbf{K}\left(\mathbf{p}_c - \mathbf{p}_r\right) = \mathbf{F}_v.~\label{eq:admittance1}
\end{equation}
The terms $\mathbf{M}$, $\mathbf{D}$ and $\mathbf{K}$ are $3\times3$ (diagonal matrices in our case) representing the mass, the damping and the stiffness acting on each axis of the world frame $W$.
The right input term to the equation is the virtual force $\mathbf{F}_v = \textbf{F}_{usr} + \textbf{F}_{rep}$ resulting from the sum of the user interaction force $\textbf{F}_{usr}$ with the obstacle repulsive force $\textbf{F}_{rep}$, which will be defined later in Section \ref{sec:Obstacle_avoidance}, while $\mathbf{F}_{usr} = \mathbf{K}_p(\mathbf{p} - \mathbf{p}_{u}) - \mathbf{K}_d(\dot{\mathbf{p}}- \dot{\mathbf{p}}_{u})$
with $\mathbf{p}$ and $\mathbf{p}_{u}$ are the drone position and the user interactive marker position respectively. These are estimated using a linear Kalman Filter (KF) for motion estimation considering a constant acceleration model which takes as input the acceleration $\ddot{\mathbf{p}}_{u}$.
The terms $\mathbf{K}_p$ and $\mathbf{K}_d$ represent respectively the proportional and the derivative terms.
Compared to the classic admittance control, the damping value $\mathbf{D}$, in the VAC definition, varies according to the direction of the input force $\mathbf{F}_v$. A variable damping enables to change the feeling of the user during the interaction since the drone is constrained to follow the direction of minimum resistance.

Starting from eq.~(\ref{eq:admittance1}) for the classic admittance controller and referring to the RRT* segment in Fig.~\ref{fig:fig2}, let $\theta = [0, \pi]$ be the angle determined by the human force direction $\mathbf{F}_v$ with respect to $\mathbf{l}_i : \theta = \cos^{-1}(\mathbf{l}_i^\top \mathbf{F}_v/||\mathbf{F}_v||)$ where $\mathbf{l}_i$
is the vector describing the direction, expressed in $W$, of the segment between two consecutive setpoints $\mathbf{P}_{S_{i-1}}$ and $\mathbf{P}_{S_{i}}$ traversed by the drone. 
Focusing on a generic segment in Fig.~\ref{fig:fig2},
$\mathbf{p}_{u}$ is projected on the segment defining the position $\mathbf{p}_{r}$, forwarded as input to the Variable Admittance Control (VAC). The sequence of reference positions $\mathbf{p}_{r}$ defines the reference trajectory $\mathbf{\Delta}_r$.
Inspired by \cite{Bazzi2022}, we define the VAC with three functions $D_k = f_k(\theta)$, with $D_k$ a diagonal element of $\mathbf{D}$, as $linear$, $squared$ and $exponential$ with
\begin{itemize}
\item $D_{k,min} = f_k(\theta = 0)$,
\item $D_{k,max} =  [f_k(\theta = \pi) \land  f_k(\theta = -\pi)]$, 
\item $f_k(\theta)$ is continuous in $[0, \pi]$ and $[0, -\pi]$.
\end{itemize}
The three damping functions $D_k = f_k(\theta)$ are integrated within the APVI mode. 
The functions modify the damping behaviour (visualized as the blue area around the drone position in Fig. \ref{fig:fig2}) depending on the pointing direction of the virtual force vector $\mathbf{F}_v$ that is forwarded as input to the admittance controller as shown in eq.~(\ref{eq:admittance1}). 

The set of commanded positions $\mathbf{p}_c$ compose the trajectory $\mathbf{\Delta}_c$, visible as a blue dashed line in Fig.~\ref{fig:fig2} (right), compliant with the direction of the force $\mathbf{F}_v$.

\subsubsection{Obstacle Perception as Force Field}
\label{sec:Obstacle_avoidance}
The FPVI modality employs the same spatial representation of the APVI case as visible in Fig.~\ref{fig:fig2} (left).
Given multiple repulsive forces per voxel $\mathbf{F}_{r_i}$ with $i\in\{0,..., N_v\}$ with $N_v$ is the total number of voxels $v_i \in V^W$ within the robot horizon $h$, it is possible to define the direction of the resulting force $\mathbf{F}_{rep}$ acting on the robot frame $B$ and expressed in the frame $W$. 
The normalized vector   
 $\mathbf{F}_{rep} = \mathbf{F}_r / \Vert\mathbf{F}_{r} \Vert $ defines the direction of the repulsive force acting on the robot, where $\mathbf{F}_r = \sum_{i=0}^{Nv} \mathbf{F}_{r_i}$ is composed by the sum of all the force vectors $\mathbf{F}_{r_i}$ generated by each of the voxel $v_i$ inside the robot horizon $h$. Each term $\mathbf{F}_{r_i}$ is therefore defined following the exponential decay behaviour in function of the euclidean distance $d$ from the voxel $v_i$ as
\begin{equation}
\Vert\mathbf{F}_{r_i}\Vert = \frac{F_s}{k}e^{(-\lambda d)} (1- e^{(h - d)}),
\label{eq:exp_decay}
\end{equation}
where $k = 1 - e^h$, $\Vert\mathbf{F}_{r_i}\Vert = F_s$ when $d = 0$ and $\lambda$ is the exponential decay constant.
Finally, at each iteration, the magnitude of the vector $\mathbf{F}_{rep}$  is defined as $\Vert\mathbf{F}_{rep} \Vert = \Vert\mathbf{F}_{r}\Vert$.

\begin{figure*}[h]
\centering
\includegraphics[width=\textwidth]{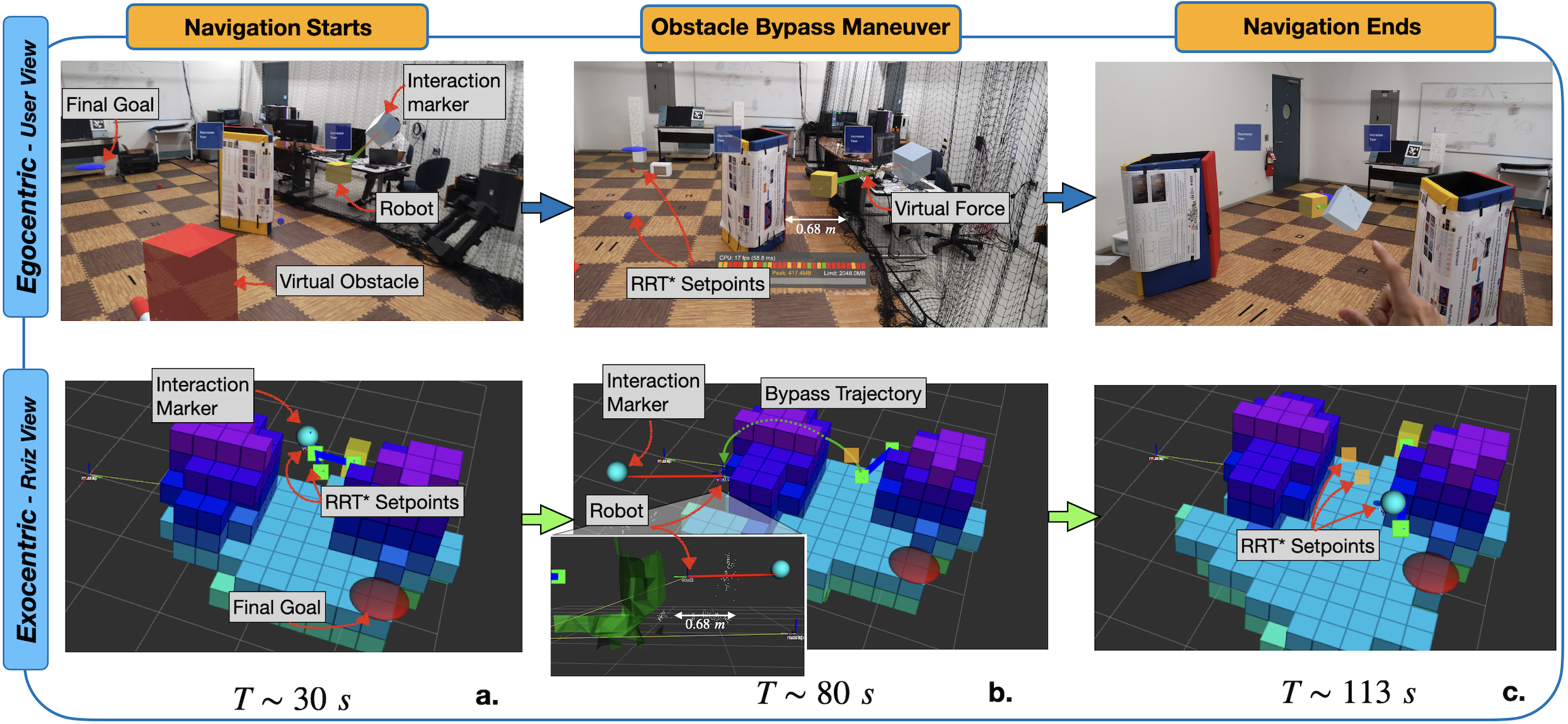}
  \caption{Ego and Exo-centric visualization of the proposed experiment captured at different salient moments. An enhanced safety layer is provided by the robot and the SAM, with the generation of a safe trajectory through the obstacles. }
  \label{fig:fig3}
  \vspace{-20pt}
\end{figure*}
\label{sec:APVI_exp}
\section{Experimental Results}~\label{sec:Experimental}
We validate the components of the proposed framework and show  their application in joint human-drone exploration tasks through multiple real-world experiments. These are
conducted at the Agile Robotics and Perception Lab (ARPL) at the New York University in the flying arena with size of $60$~\si{m^2} as well as in an adjacent room to show the fully capabilities of the tele-immersive framework in collaborative tasks, as proposed in Fig.~\ref{fig:fig1}.b and Fig.~\ref{fig:fig1}.c. 
 
The quadrotor platform used in the experiments is equipped with a $\text{Qualcomm}^{\circledR}\text{Snapdragon}^{\text{TM}}$ board for onboard computing where an embedded stereo camera is used for a disparity representation, obtaining the point cloud ${{P_c}_{robot}}^B$. Our algorithm provides the state of the robot (pose and perception output) directly in the $W$ frame via the transformation $\mathbf{T}_{B}^{W}$. In addition, a mutual co-localization mechanism between the MR device and the aerial robots can be used\cite{azure_spatial_anchors_ros}. Both the robot and the AHRMD use onboard self-localization mechanism.
The Mixed Reality framework is implemented in C$\#$ and it is executed on the Microsoft$^{\circledR}$ Hololens 2.0.
Combining different features, the proposed framework offers four types of deployment modalities
 \begin{itemize}
     \item \textbf{Full Simulation (FS)} The complete interaction is based on ARPL custom quadrotor simulator and visualized in ROS RViz.
     \item \textbf{Mixed Simulation (MS)}: It is the same as \textbf{FS}, but the user can interact with the robot using the MR headest.  
     \item \textbf{Real Interaction (RI)}: The user employs ROS Rviz to send commands to the real flying drone. In this case, the mapping and navigation functionalities are enabled.
     \item \textbf{Real Interaction in MR (RIMR)}: In this case the framework is fully enabled.
 \end{itemize}
We provide an experimental overview of the FPVI and APVI interaction modalities proposed in Section \ref{sec:PVH interaction} and shown in Fig.~\ref{fig:fig2}.
These modalities are tested in a lab environment by four individuals, all of whom provided positive feedback regarding the projected amount of visual information and the ease of interaction with an aerial robot. They consistently achieved the assigned task to reach a target in a densely populated obstacle environment, relying on both the robot perception feedback represented by the computed planner path and the obstacle repulsive force. A visualization of two of the possible tasks assigned to the subjects are described in the following sections.

\subsection{VAC with RRT* exploration and Mapping}
\label{sec:VAC_RRT_experiments}
\begin{figure}[!t]
\centering
\includegraphics[width=\columnwidth]{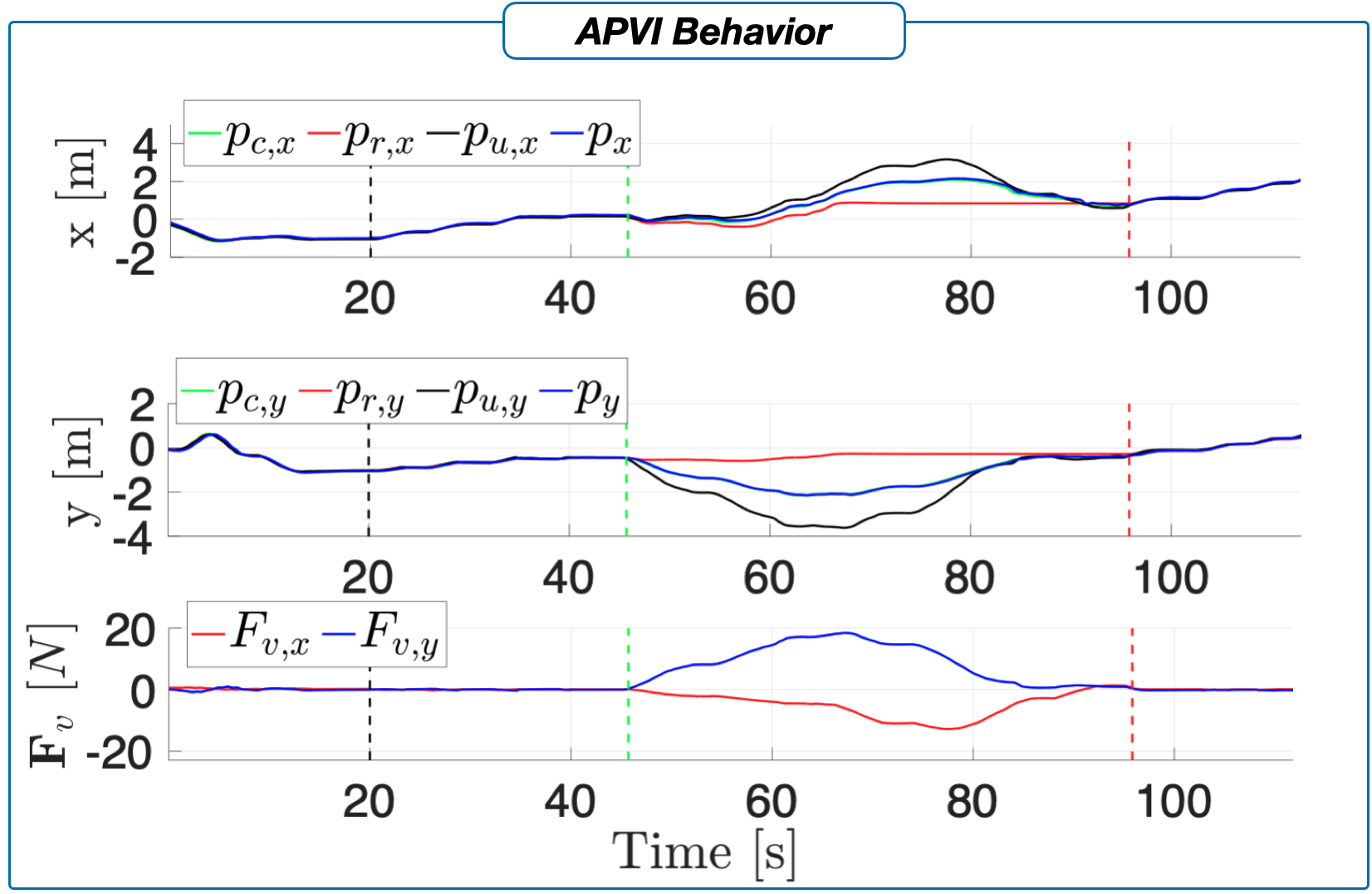}
  \caption{VAC behavior during the APVI interaction.}
  \label{fig:fig4}
  \vspace{-10pt}
\end{figure}

The interaction behaviour during the APVI mode is represented in  Fig.~\ref{fig:fig3} showing a sequence of frames with an Egocentric and Exocentric views at different time steps during the same experiment whose results are in Fig.~\ref{fig:fig4}. 

During the experiments the height of the robot is kept constant at $p_{r,z} = 0.8~\si{m}$.
The black dashed line at time instant $t_{APVI} = 19.5$~\si{s}, in Fig.~\ref{fig:fig4}, represents the switches from (FPVI) to (APVI) mode which happens when the user places a final goal to the robot to reach, denoted as  $\mathbf{G}^W$ and visualized in Fig.~\ref{fig:fig3} with a red circle. 
Once activated, the RRT* obstacle aware planning algorithm, coupled with the VAC initialized with $M_k=2.4~\si{Kg}$, $K_k = 20~\si{Kg \cdot s^{-2}}$ and $1~\si{Kg \cdot s^{-1}}\leq D_k \leq 70~\si{Kg \cdot s^{-1}}$, where the subscript $k$ represents the diagonal element of $\mathbf{M}$, $\mathbf{K}$, and $\mathbf{D}$ respectively, constraints the drone to follow the proposed path. 
Nonetheless, the user can still pull away the robot from the initial trajectory despite this safety condition, applying an increasing virtual Force $\mathbf{F}_v$ as input to the system, in order to define a maneuver to bypass virtual and real obstacles. 
This makes the robot movement compliant to the user intentions but still constrained to follow the reference trajectory $\mathbf{\Delta}_r$, relying on the planner proposed path. 
The maneuver and the human-drone assisted navigation are visible in detail also in the sequence of frames in Fig.~\ref{fig:fig3}. 

To show the system's compliance to the human intentions, in Fig.~\ref{fig:fig3}.b, we observe that the user's action can pull the robot away from the RRT* planned path and drive it along a new trajectory passing through a narrow gap measured in only $0.68~\si{m}$ (schematized as a green arrow).
The estimated obstacle Truncated Signed Distance Field (TSDF) \cite{Curless1996,werner2014}, represented by the green mesh $M_{robot}$, visualizes the detected surface of the obstacle as perceived by the drone sensors during the bypass maneuver.
Once this maneuver terminates, in Fig.~\ref{fig:fig3}.c, the RRT* restarts its expansion towards $\mathbf{G}^W$, only when the drone is attracted back to the last generated setpoint.
Moreover, in Figs.~\ref{fig:fig3}.a, \ref{fig:fig3}.b, and \ref{fig:fig3}.c the interaction marker (white cube), the drone camouflage (orange cube) and the generated force (colorful string) are easily identifiable by the user.
The generated trajectory depending on the Force $\mathbf{F}_v$ is shown in Fig. \ref{fig:fig4}. The Root Mean Squared Error between $\mathbf{\Delta}_r$ and $\mathbf{\Delta}_c$ during the FPVI and APVI modalities is respectively of $0.0373$~\si{m} and $2.0142$~\si{m}.
The higher RMSE value obtained during the APVI phase is due to the obstacle bypass maneuver, where the commanded trajectory is compliant with the user intentions rather than following the original desired path.

\subsection{FPVI with Obstacle Force Field}
\label{sec:VAC_FPVI_obstacle}
\begin{figure}[t!]
\includegraphics[width=\columnwidth, keepaspectratio]{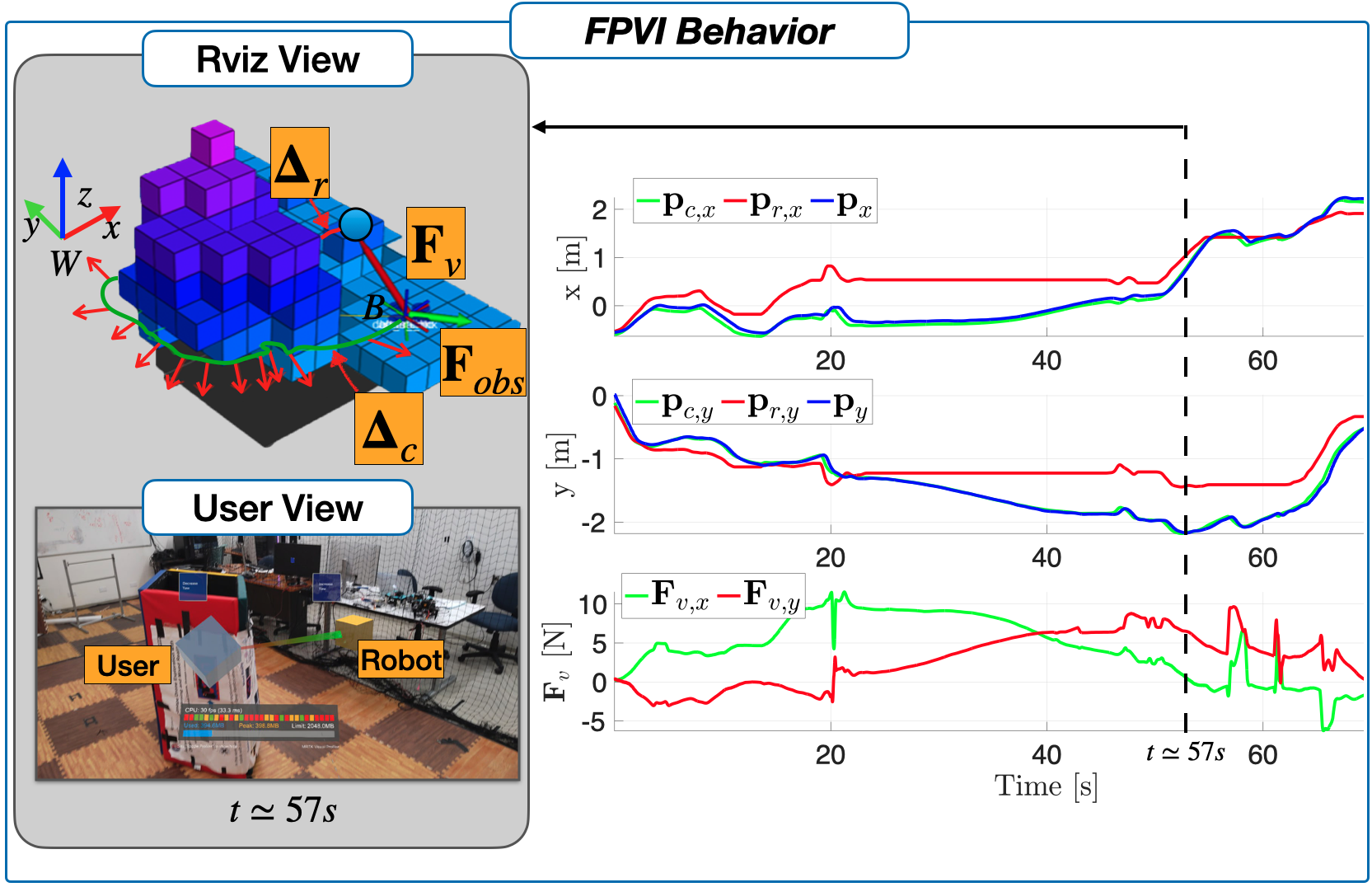}
  \caption{Quantitative visualization of the obstacle safety layer acting on the system visualized in User and Rviz view}
  \label{fig:fig5}
  \vspace{-15pt}
\end{figure}

We propose a second experiment which provides a deeper insight of the behavior of the system when subject to the action of the obstacle generated force field, defined in Section~\ref{sec:Obstacle_avoidance} when the system is set to the FPVI mode.
Along the overall experiment, the gains of the admittance controller, except for $D_k = 20~\si{Kg \cdot s^{-1}}$, are not modified with respect the previous scenario shown in Section~\ref{sec:APVI_exp}. The exponential decay behaviour of the force $\mathbf{F}_{rep}$ is defined respectively with $Fs = 8~\si{N}$, $\lambda = 1$ and horizon $h = 1.5~\si{m}$. 
Referring to Section~\ref{eq:exp_decay} and following the exponential decay physical behavior, the magnitude of the force $\Vert \mathbf{F}_{rep} \Vert = 0$ when $d = h$ and  $\Vert \mathbf{F}_{rep} \Vert = F_s$ when $d = 0$.
The results of this experiment are provided in Fig.~\ref{fig:fig5}, with a detailed Ego ("User view") and Exo-centric ("Rviz view") representations of the interaction captured at time $t = 57 ~\si{s}$.
As visible from the location of the white interaction cube, visualized as a sphere in the map, the user tries to impose an unsafe desired trajectory $\mathbf{\Delta}_r$ passing through an obstacle. As the robot perceives the environment a repulsive force $\mathbf{F}_{rep}$, defined as in \ref{sec:Obstacle_avoidance}, is generated and visualized in the Exo-centric representation in Fig.~\ref{fig:fig5} as a green arrow. The user pulling force and the obstacle repulsive force are forwarded as input to the admittance controller which generates new safer commanded trajectory $\mathbf{\Delta}_c$, visualized in Fig.~\ref{fig:fig5} as a green line. 
The complete evolution of the trajectories $\mathbf{\Delta}_c$, $\mathbf{\Delta}_r$ and the robot actual sequence of positions $\mathbf{p}$, are represented in the plot in Fig.~\ref{fig:fig5} where the components of the trajectories along the $x,~y$ axis are visualized. 
The "User View" in Fig.~\ref{fig:fig5} provides an enhanced visualization of the human robot interaction at the time of the representation in the "Rviz view" as seen through the head mounted device.

\subsection{Results Discussion}

The primary objective of this work is to validate the technical modules of proposed framework.
The presented results validate each component and the overall framework's functionality, showcasing potential for natural and intuitive interaction beyond traditional tele-operation and basic MR solutions. Tests in a cluttered environment show that safety measures introduced through the planner algorithm (Section \ref{sec:VAC_RRT_experiments})  and obstacle force field (Section \ref{sec:VAC_FPVI_obstacle}) enable secure, gesture-based interaction with the drone, adaptable to user needs. Tests performed by multiple users further validate the methodologies and the benefits of the proposed solution. This approach serves as a flexible building block, open to incorporation with other modalities for enhanced bi-directional data interpretation and interaction.

\section{Conclusion}~\label{sec:Conclusion}
In this paper, we presented a  tele-immersive human-robot collaborative framework. The key features include a novel bi-directional spatial awareness approach that provides seamless human-robot spatial awareness through MR and a multi-modal virtual-physical interaction that further enhances this collaboration experience. We open-source the framework with the goal to promote research in this area and incorporate feedback from the community.

Despite four users already validated the usability of the system in the lab, in the future, we expect to conduct a complete user case study, leveraging the NASA Task Load Index \cite{Nasahart1988development}, subjective questions and quantitative data collected on an inspection target reaching based task, in order to have a complete performance evaluation on multiple users. This will help to further validate the presented approaches and results.
Specifically, we would like to compare the users' feelings during a 2D based interaction and full MR experience. We believe the results will guarantee an easier manipulation and a continuity in the interaction between the robot and non expert pilots when both agents share the same environment.
Finally, we aim to extend the proposed framework by allowing the incorporation of heterogeneous teams consisting of multiple users and robots.


\bibliographystyle{IEEEtran}
\bibliography{ICRA2024}

\end{document}